\documentclass[pdflatex,sn-mathphys-num]{sn-jnl}

\usepackage{threeparttable}%
\usepackage{graphicx}%
\usepackage{multirow}%
\usepackage{amsmath,amssymb,amsfonts}%
\usepackage{amsthm}%
\usepackage{mathrsfs}%
\usepackage[title]{appendix}%
\usepackage{xcolor}%
\usepackage{textcomp}%
\usepackage{manyfoot}%
\usepackage{booktabs}%
\usepackage{tabularx}
\usepackage{algorithm}%
\usepackage{algorithmicx}%
\usepackage{algpseudocode}%
\usepackage{listings}%
\usepackage{makecell}
\usepackage{array}
\usepackage{hyperref} 
\usepackage{url} 
\usepackage{mathrsfs}
\newcolumntype{Y}{>{\centering\arraybackslash}X}
\usepackage{caption} 

\theoremstyle{thmstyleone}%
\theoremstyle{thmstyletwo}%

\theoremstyle{thmstylethree}%

\begin{document}


\title{One-shot synthesis of rare gastrointestinal lesions improves diagnostic accuracy and clinical training}



\author[1,2]{\fnm{Jia} \sur{Yu}}
\equalcont{These authors contributed equally to this work.}

\author[4,5]{\fnm{Yan} \sur{Zhu}}
\equalcont{These authors contributed equally to this work.}

\author[4,5]{\fnm{Peiyao} \sur{Fu}}

\equalcont{These authors contributed equally to this work.}

\author[4]{\fnm{Tianyi} \sur{Chen}}

\author[1,2]{\fnm{Zhihua} \sur{Wang}}

\author[1,2]{\fnm{Fei} \sur{Wu}}

\author[4,5]{\fnm{Quanlin} \sur{Li}}

\author*[4,5]{\fnm{Pinghong} \sur{Zhou}}\email{zhou.pinghong@zs-hospital.sh.cn}
\author*[3,4,7]{\fnm{Shuo} \sur{Wang}}\email{shuowang@fudan.edu.cn}

\author*[6,7]{\fnm{Xian} \sur{Yang}}\email{xian.yang@manchester.ac.uk}

\affil[1]{%
  \orgname{Zhejiang University},
  \orgaddress{\city{Hangzhou}, \country{China}}}

\affil[2]{%
  \orgname{Shanghai Institute for Advanced Study of Zhejiang University},
  \orgaddress{\city{Shanghai}, \country{China}}}

\affil[3]{%
  \orgname{Digital Medical Research Center, School of Basic Medical Sciences, Fudan University},
  \orgaddress{\city{Shanghai}, \country{China}}}

\affil[4]{%
  \orgname{Shanghai Collaborative Innovation Center of Endoscopy},
  \orgaddress{\city{Shanghai}, \country{China}}}

\affil[5]{%
  \orgdiv{Endoscopy Centre and Endoscopy Research Institute},
  \orgname{Zhongshan Hospital, Fudan University},
  \orgaddress{\city{Shanghai}, \country{China}}}

\affil[6]{%
  \orgname{Alliance Manchester Business School, The University of Manchester},
  \orgaddress{\city{Manchester}, \country{UK}}}

\affil[7]{%
  \orgname{Data Science Institute, Imperial College London},
  \orgaddress{\city{London}, \country{UK}}}



\abstract{Rare gastrointestinal lesions are infrequently encountered in routine endoscopy, restricting the data available for developing reliable artificial intelligence (AI) models and training novice clinicians. Here we present EndoRare, a one-shot, retraining-free generative framework that synthesizes diverse, high-fidelity lesion exemplars from a single reference image. By leveraging language-guided concept disentanglement, EndoRare separates pathognomonic lesion features from non-diagnostic attributes, encoding the former into a learnable prototype embedding while varying the latter to ensure diversity. We validated the framework across four rare pathologies (calcifying fibrous tumor, juvenile polyposis syndrome, familial adenomatous polyposis, and Peutz–Jeghers syndrome). Synthetic images were judged clinically plausible by experts and, when used for data augmentation, significantly enhanced downstream AI classifiers, improving the true positive rate at low false-positive rates. Crucially, a blinded reader study demonstrated that novice endoscopists exposed to EndoRare-generated cases achieved a 0.400 increase in recall and a 0.267 increase in precision. These results establish a practical, data-efficient pathway to bridge the rare-disease gap in both computer-aided diagnostics and clinical education.}




\maketitle

\section{Introduction}\label{intro}
Rare diseases affect over 300 million people worldwide, yet timely and accurate diagnosis remains challenging. This challenge is especially acute for rare gastrointestinal lesions in endoscopic practice, where many entities are documented by only a single image or a handful of cases~\cite{schaefer2020use,ktena2024generative,zhang2025generative}. Because endoscopic diagnosis is a dynamic, real-time process that relies on continuous multi-angle visual inspection and clinical experience rather than direct instrument readouts, limited exposure makes it difficult for clinicians to build reliable recognition of rare phenotypes.
The same scarcity also leaves data-driven artificial intelligence (AI) models with too few examples to learn robust diagnostic patterns. Together, these factors increase the risk of missed or incorrect diagnoses~\cite{vetro2015rare,hawkes2012diagnosis,urban2018deep,esteva2019guide}.

Conventional augmentation is often insufficient for rare endoscopic lesions because it cannot reproduce the clinically meaningful appearance changes that shape diagnosis, including variations in lighting, viewing geometry, mucosal background, lesion scale, and fine surface texture. As a result, augmented images may increase quantity without providing the diversity that supports reliable recognition.
Recent text-guided generative models, such as MINIM~\cite{wang2024self}, can synthesize high-fidelity medical images across multiple organs and modalities and have been explored for improving downstream AI-assisted diagnosis via data augmentation. In endoscopic imaging, CCIS-Diff~\cite{xie2024ccis} generates controllable colonoscopy images using a diffusion-model prior. However, these approaches are difficult to apply to rare or sparsely documented lesions because they typically require substantial paired text and image data for effective adaptation, a requirement that is rarely met in extreme low-data settings.

To adapt generative models to a new lesion type when only a few examples exist, prior few-shot strategies commonly rely on concept-specific optimization, such as fine-tuning a pre-trained diffusion model or jointly optimizing prompts and model parameters~\cite{gal2022image,ruiz2023dreambooth,kumari2023multi}. In practice, maintaining a separate adapted model for each rare phenotype does not scale~\cite{kazerouni2023diffusion}, and learning from a single reference image often overfits and reduces clinically acceptable variation. Training-free seed selection methods (e.g., SeedSelect~\cite{samuel2024generating}) can improve sample diversity, but their gains are often brittle across different prompts and clinical contexts. Prompt engineering approaches such as DATUM~\cite{benigmim2023one} may further increase diversity, yet they risk altering diagnostically critical morphology, which limits their suitability for rare disease synthesis.

\begin{figure}[b]
    \centering
    \includegraphics[width=1.0\linewidth]{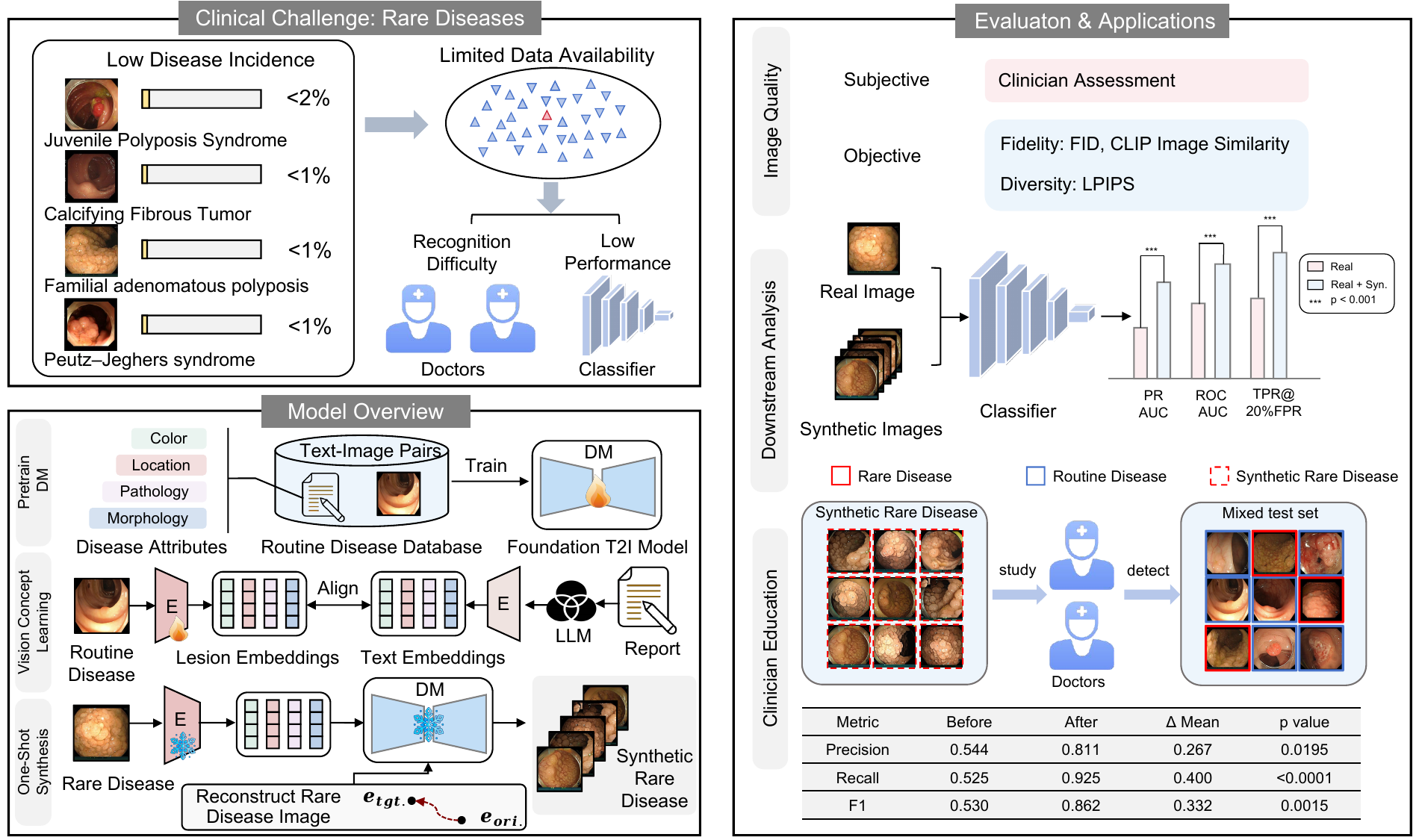}
    \caption{\textbf{Overview of the EndoRare framework.}
    \textbf{a}, Clinical motivation. Long-tail rarity in gastrointestinal diseases causes extreme data scarcity, limiting both clinician exposure and AI training.
    \textbf{b}, Framework. Three stages transfer knowledge from routine to rare lesions: diffusion pretraining on routine image--text pairs, attribute alignment between images and text, and controllable synthesis of rare cases using learned embeddings and disentangled attributes.
    \textbf{c}, Evaluation. We assess (1) image quality and clinical faithfulness, (2) gains in AI-assisted diagnosis when training data are augmented with synthetic rare cases, and (3) clinical relevance via improved diagnostic performance of novice trainees.}
    \label{fig:overview}
\end{figure}

Therefore, there is a pressing need for a cost-effective framework that can robustly adapt to and faithfully synthesize target rare lesions.
We propose \textbf{EndoRare}, a retraining-free one-shot framework for endoscopic rare-lesion synthesis. {EndoRare} pretrains a medical-domain diffusion prior and integrates language-informed concept disentanglement to separate and control clinically meaningful attributes. During personalization, it freezes the diffusion backbone and learns (i) a compact prototype embedding from a single exemplar and (ii) encoder-derived, language-aligned visual concepts. These representations steer the denoising trajectory to produce diverse, clinically credible samples without any additional fine-tuning at inference.

Empirically, {EndoRare} achieves state-of-the-art performance on endoscopic datasets. In blinded clinician assessments of realism and class faithfulness, it consistently outperforms prior generative baselines. When used for data augmentation, {EndoRare}-generated samples lead to reliable improvements in AI-assisted diagnosis of rare lesions compared with conventional augmentation. To facilitate clinical adoption, exposure to EndoRare-generated cases also enhances trainees’ recognition of rare diseases, highlighting both translational and educational impact.

\section{Results}\label{results}

\subsection{Overview of EndoRare framework}

Fig.~\ref{fig:overview} summarizes the EndoRare framework and evaluation design. EndoRare addresses the long-tail rarity in endoscopy that limits both clinical exposure and data-driven learning (Fig.~\ref{fig:overview}a) through a three-stage pipeline (Fig.~\ref{fig:overview}b): pretraining a knowledge-informed diffusion model on routine image--text pairs, learning an attribute-aligned image encoder, and deploying a controllable generator that combines a lesion-specific embedding with attributes extracted from real rare cases. Methodological details are provided in Section~\ref{method}.

We evaluate EndoRare along three axes (Fig.~\ref{fig:overview}c) on four rare entities, CFT~\cite{turbiville2020calcifying}, JPS~\cite{brosens2011juvenile}, FAP~\cite{beech2001familial}, and PJS~\cite{latchford2022gastrointestinal} (Supplementary Table~B1): (i) image fidelity and clinical faithfulness using quantitative metrics and blinded clinician ratings; (ii) downstream utility by training rare-lesion diagnostic models with EndoRare-augmented data against conventional augmentation and recent generative baselines; and (iii) clinical relevance via a within-subject pre--post reader study: novices were first trained using only one real anchor image per entity and then evaluated on a fixed mixed image set containing both rare and non-rare cases, where their task was to detect and identify rare lesions; they were subsequently retrained with additional {EndoRare}-generated exemplars and re-evaluated on the same mixed set.

\begin{figure}[!]
  \centering
  \includegraphics[width=\textwidth]{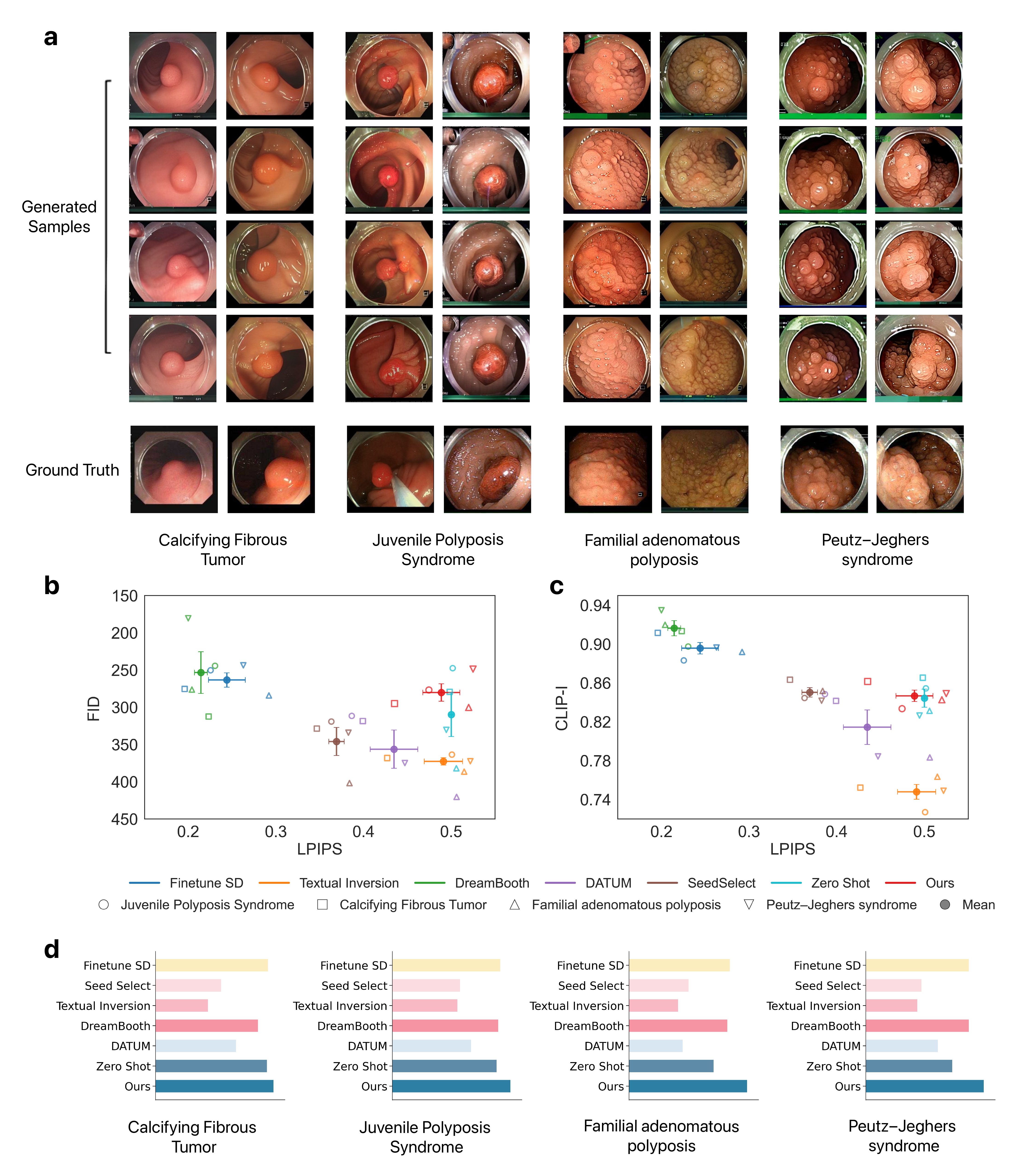}
  \caption{\textbf{Visual realism and class faithfulness.}
  \textbf{a}, Representative generated samples (rows 1–4) alongside ground-truth exemplars (bottom row) for four diseases.
  \textbf{b}, Image fidelity vs. diversity: class-wise points and per-method means (circles with s.e.m. bars) in the FID–LPIPS plane. The y-axis is inverted (lower FID is better) and higher LPIPS indicates greater diversity; better performance is towards the \textbf{upper-right}.
  \textbf{c}, Image–image consistency vs. diversity: CLIP-I–LPIPS scatter with per-method means (s.e.m. bars). Higher CLIP-I and higher LPIPS are desirable; better performance is towards the \textbf{upper-right}.
  \textbf{d}, Clinical evaluation: expert gastroenterologists score generated images on a 3-point scale (1\,=\,unrealistic; 3\,=\,realistic and class-faithful without exact replication).}
  \label{fig:case_study}
\end{figure}

\begin{figure}[!b]
  \centering
  \includegraphics[width=\textwidth]{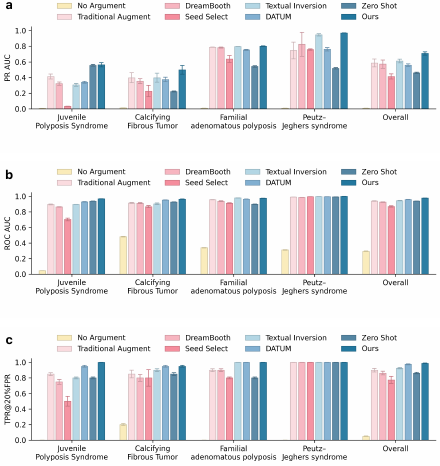}
  \caption{\textbf{Classification performance with generated data.}
  Eight strategies are compared across four rare-lesion categories and their macro-average ({Overall}). 
  \textbf{a}, Precision–recall area under the curve (PR-AUC).
  \textbf{b}, Receiver operating characteristic area under the curve (ROC-AUC).
  \textbf{c}, \(\mathrm{TPR}@20\%\ \mathrm{FPR}\) (true positive rate at the operating point with the false-positive rate fixed at \(20\%\)).
  Bars in \textbf{a–c} denote means across runs; error bars indicate standard deviation.
}
  \label{fig:classification}
\end{figure}

\begin{figure}[!b]
  \centering
  \includegraphics[width=1.0\linewidth]{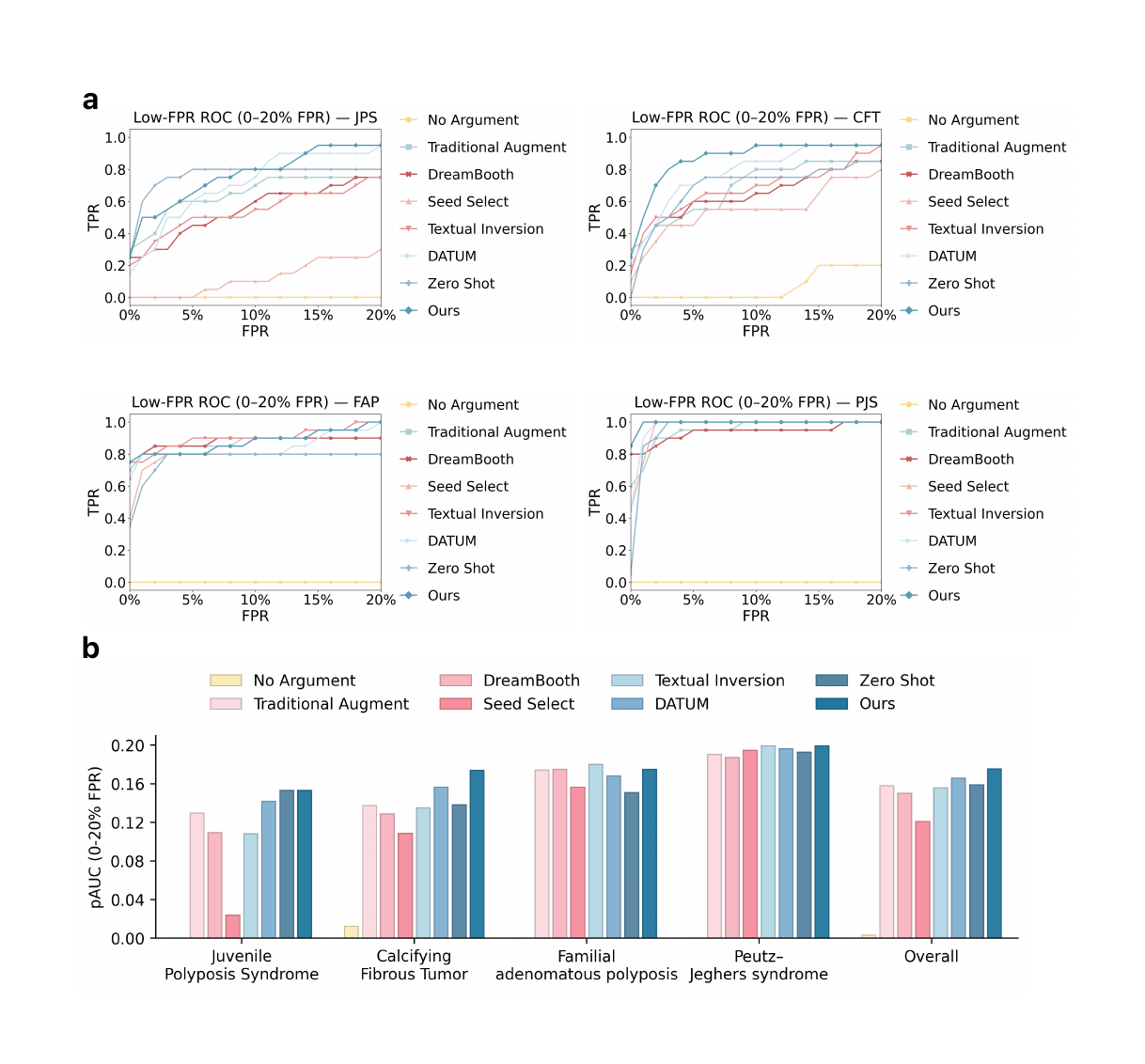}
  \caption{\textbf{Low-FPR detection performance (0–20\% FPR).}
  \textbf{a}, Per-entity ROC curves in the clinically relevant low-FPR range (JPS, CFT, FAP, PJS). 
  \textbf{b}, Partial AUC (pAUC) computed over the same 0–20\% FPR interval, summarized per entity and as a macro-average (Overall).
  Higher pAUC indicates greater cumulative sensitivity under low-false-positive operating constraints.}
  \label{fig:lowfpr_roc_pauc}
\end{figure}


\subsection{High-fidelity and diverse synthesis of rare lesions}
We first evaluated the visual realism and clinical relevance of the synthesized images. As shown in Fig.~\ref{fig:case_study}a, {EndoRare} successfully reproduces critical endoscopic photometrics, such as mucosal sheen, specular highlights, and natural vignetting, while preserving the hallmark morphology of specific rare entities (e.g., solitary nodules for CFT/JPS versus carpet-like clusters for FAP/PJS).
 The variations in viewpoint, scale, and luminal background demonstrate that the model achieves substantial intra-class diversity without compromising class semantics.

To quantify this performance, we analyzed the trade-off between image fidelity and diversity (Fig.~\ref{fig:case_study}b). While DreamBooth achieved the lowest Fréchet Inception Distance (FID~\cite{heusel2017gans}: $253.43 \pm 28.02$), it suffered from limited diversity (LPIPS~\cite{zhang2018unreasonable}: $0.214 \pm 0.007$). Conversely, the Zero Shot baseline maximized diversity ($0.500 \pm 0.002$) but at the cost of significantly degraded fidelity (FID: $309.82 \pm 29.49$). {EndoRare} struck a favorable balance between these regimes, maintaining competitive fidelity (FID: $280.26 \pm 11.64$) while achieving high diversity (LPIPS: $0.489 \pm 0.021$). In terms of semantic consistency (Fig.~\ref{fig:case_study}c), {EndoRare} (CLIP-I~\cite{radford2021learning}: $0.847 \pm 0.006$) remained competitive with other generative baselines, ensuring that the synthesized diversity remains faithful to the text prompts.

Crucially, blinded assessments by expert gastroenterologists corroborated these objective metrics (Fig.~\ref{fig:case_study}d). In a study involving 700 randomly sampled images evaluated on a 3-point clinical scale, {EndoRare} achieved the highest overall rating (2.32), significantly outperforming fine-tuned Stable Diffusion (2.09) and DreamBooth (2.01), as well as other baselines such as Zero Shot (1.90) and DATUM (1.40). Experts noted that {EndoRare} not only generated realistic textures but also accurately preserved characteristic lesion features without mere replication, yielding substantial inter-rater agreement (Fleiss' $\kappa = 0.59$). Detailed breakdowns of these metrics are provided in Supplementary Tables~B2–B6 and Figs.~A6–A12.

\subsection{Improved automated diagnosis via synthetic augmentation}

We evaluated the downstream utility of {EndoRare} by benchmarking a rare-lesion classifier trained with our augmented data against traditional augmentation and recent generative baselines. As summarized in Fig.~\ref{fig:classification}a–c, incorporating {EndoRare} significantly enhances overall diagnostic performance. Specifically, our method raised the macro-average PR-AUC to 0.708, surpassing the strongest baseline (Textual Inversion, 0.612) by a substantial margin (+0.096, $p<0.001$). Similarly, {EndoRare} achieved the highest macro-average ROC-AUC of 0.978 and improved sensitivity at a clinically relevant operating point (TPR@20\% FPR = 0.988), offering an absolute gain of 8.8\% over traditional augmentation ($p<0.001$).

This performance advantage extends across individual disease categories. {EndoRare} yielded the most pronounced improvements in PR for challenging entities such as CFT (0.498 vs.\ 0.398 for Textual Inversion; $p<0.001$). In terms of ROC-AUC, performance approached ceiling levels for several categories; notably, {EndoRare} led significantly on JPS ($p<0.001$), CFT ($p<0.05$), and PJS ($p<0.001$). While performance on FAP was saturated across methods, {EndoRare} remained highly competitive (0.975), comparable to the top-performing baselines. Saturation effects were also observed for TPR@20\% FPR, where {EndoRare} attained perfect scores (1.0) on JPS and FAP and tied for top performance on CFT (0.95).

Given that minimizing false positives is critical for reducing procedural burden in endoscopy, we further analyzed performance within the constrained low-FPR window (0--20\%, Fig.~\ref{fig:lowfpr_roc_pauc}). {EndoRare} yielded the highest macro partial AUC (pAUC) of 0.175, exceeding the strongest baseline (DATUM) by 5.8\%. Entity-specific analysis confirmed that the largest gains occurred in CFT ($\Delta=+0.017$), while PJS reached saturation (pAUC = 0.199). Although JPS performance was tied with Zero Shot in terms of pAUC, {EndoRare} demonstrated superior sensitivity at the 20\% FPR operating point (0.95 vs.\ 0.80). Collectively, these results indicate that {EndoRare}-augmented training maintains higher cumulative sensitivity within the decision regions most relevant to clinical screening, preserving robustness across diverse rare entities.

\begin{figure}[!t]
    \centering
    \includegraphics[width=\linewidth]{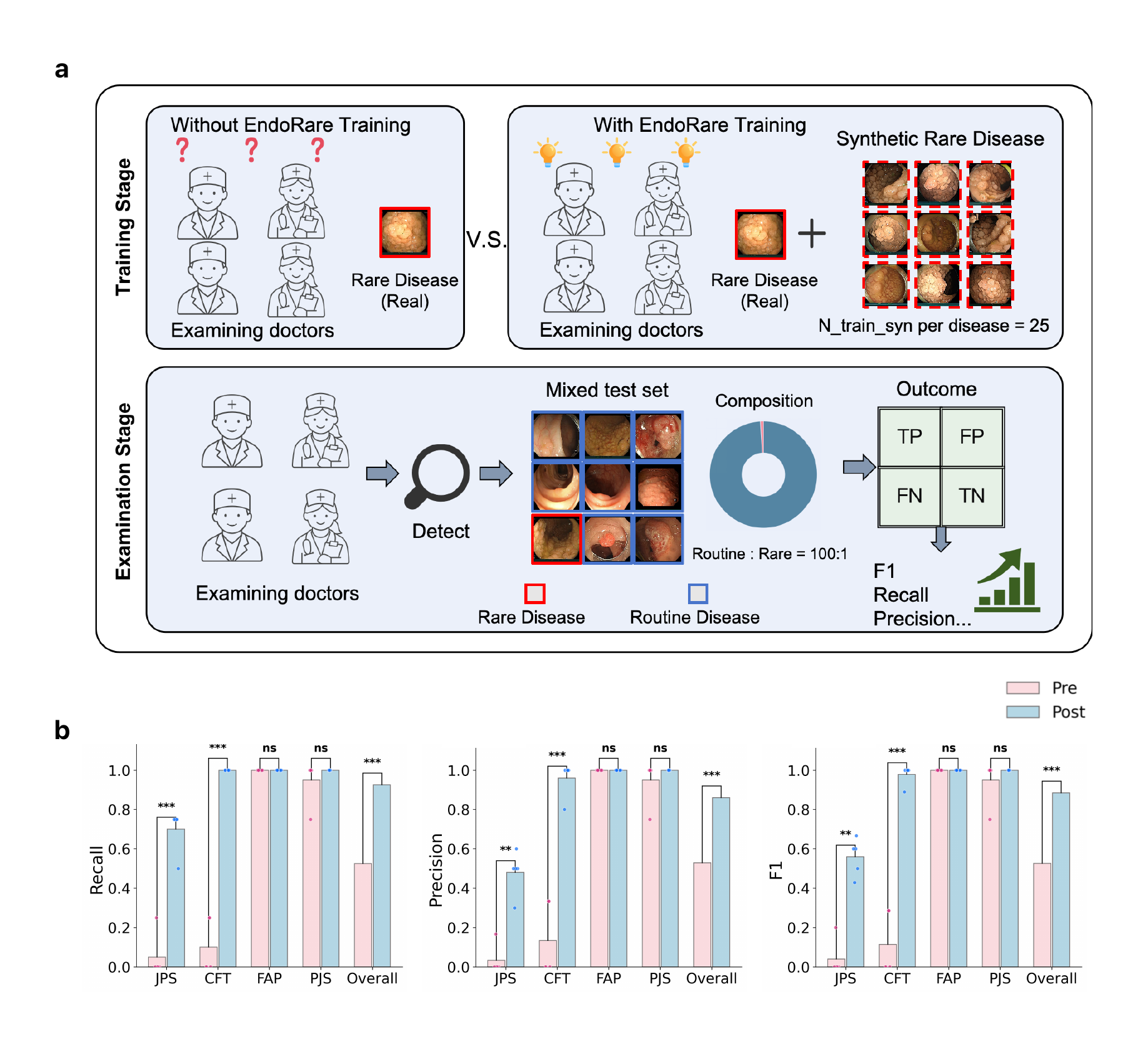}
    \caption{%
    \textbf{Impact of synthetic exposure on novice diagnostic performance.}
    \textbf{a}, Study workflow. Conventional training suffers from data scarcity (top), limiting the ability of novices to recognize rare patterns. In contrast, {EndoRare} enrichment (bottom) provides diverse synthetic exemplars to bridge this experience gap.
    \textbf{b}, Quantitative assessment. Pairwise comparison of diagnostic metrics (Recall, Precision, F1) before (\emph{Pre}) and after (\emph{Post}) synthetic training. Significant improvements are observed for JPS, CFT, and overall performance, while FAP and PJS exhibit ceiling effects due to high baseline distinctiveness.}
    \label{fig:reader_training}
\end{figure}

\subsection{Synthetic exposure improves novice diagnostic performance}

\begin{figure}[!]
  \captionsetup{font=small, skip=4pt}
  \centering
  \begin{minipage}{\textwidth}
    \captionsetup{type=table}
    \small
    \setlength{\tabcolsep}{1.5pt}
    \renewcommand{\arraystretch}{1.15}
    \caption{Per-endoscopist performance before (Pre) and after (Post) training across four entities.}
    \label{tab:reader_training_by_disease}
    \begin{tabularx}{\textwidth}{l*{12}{Y}}
      \toprule
      \multirow{2}{*}{\textbf{Case}} &
      \multicolumn{3}{c}{\textbf{\makecell{JPS}}} &
      \multicolumn{3}{c}{\textbf{\makecell{CFT}}} &
      \multicolumn{3}{c}{\textbf{\makecell{FAP}}} &
      \multicolumn{3}{c}{\textbf{\makecell{PJS}}} \\
      \cmidrule(lr){2-4}\cmidrule(lr){5-7}\cmidrule(lr){8-10}\cmidrule(lr){11-13}
      & \textbf{Prec.\ (\%)} & \textbf{Rec.\ (\%)} & \textbf{F1 (\%)} 
      & \textbf{Prec.\ (\%)} & \textbf{Rec.\ (\%)} & \textbf{F1 (\%)} 
      & \textbf{Prec.\ (\%)} & \textbf{Rec.\ (\%)} & \textbf{F1 (\%)} 
      & \textbf{Prec.\ (\%)} & \textbf{Rec.\ (\%)} & \textbf{F1 (\%)} \\
      \midrule
      (1, {Pre})  & 16.7 & 25.0 & 20.0 & 0.0 & 0.0 & 0.0 & 100.0 & 100.0 & 100.0 & 100.0 & 100.0 & 100.0 \\
      (1, {Post}) & 30.0 & 75.0 & 42.9 & 100.0 & 100.0 & 100.0 & 100.0 & 100.0 & 100.0 & 100.0 & 100.0 & 100.0 \\
      \cmidrule(lr){1-13}
      (2, {Pre})  & 0.0 & 0.0 & 0.0 & 0.0 & 0.0 & 0.0 & 100.0 & 100.0 & 100.0 & 75.0 & 75.0 & 75.0 \\
      (2, {Post}) & 50.0 & 50.0 & 50.0 & 80.0 & 100.0 & 88.9 & 100.0 & 100.0 & 100.0 & 100.0 & 100.0 & 100.0 \\
      \cmidrule(lr){1-13}
      (3, {Pre})  & 0.0 & 0.0 & 0.0 & 0.0 & 0.0 & 0.0 & 100.0 & 100.0 & 100.0 & 100.0 & 100.0 & 100.0 \\
      (3, {Post}) & 60.0 & 75.0 & 66.7 & 100.0 & 100.0 & 100.0 & 100.0 & 100.0 & 100.0 & 100.0 & 100.0 & 100.0 \\
      \cmidrule(lr){1-13}
      (4, {Pre})  & 0.0 & 0.0 & 0.0 & 33.3 & 25.0 & 28.6 & 100.0 & 100.0 & 100.0 & 100.0 & 100.0 & 100.0 \\
      (4, {Post}) & 50.0 & 75.0 & 60.0 & 100.0 & 100.0 & 100.0 & 100.0 & 100.0 & 100.0 & 100.0 & 100.0 & 100.0 \\
      \cmidrule(lr){1-13}
      (5, {Pre})  & 0.0 & 0.0 & 0.0 & 33.3 & 25.0 & 28.6 & 100.0 & 100.0 & 100.0 & 100.0 & 100.0 & 100.0 \\
      (5, {Post}) & 50.0 & 75.0 & 60.0 & 100.0 & 100.0 & 100.0 & 100.0 & 100.0 & 100.0 & 100.0 & 100.0 & 100.0 \\
      \midrule
      Mean (Pre)          & 3.3 & 5.0 & 4.0 & 13.3 & 10.0 & 11.4 & 100.0 & 100.0 & 100.0 & 95.0 & 95.0 & 95.0 \\
      Mean (Post)         & 48.0 & 70.0 & 55.9 & 96.0 & 100.0 & 97.8 & 100.0 & 100.0 & 100.0 & 100.0 & 100.0 & 100.0 \\
      $\Delta$ Mean       & 44.7 & 65.0 & 51.9 & 82.7 & 90.0 & 86.4 & 0.0 & 0.0 & 0.0 & 5.0 & 5.0 & 5.0 \\
      $p$ (paired $t$)    & 0.0052 & 0.0004 & 0.0026 & 0.0004 & 0.0001 & 0.0002 & NA & NA & NA  & 0.3739 & 0.3739 & 0.3739 \\
      \bottomrule
    \end{tabularx}
  \end{minipage}

  \vspace{6pt}

  \begin{minipage}{\textwidth}
    \centering
    \includegraphics[width=1.0\linewidth]{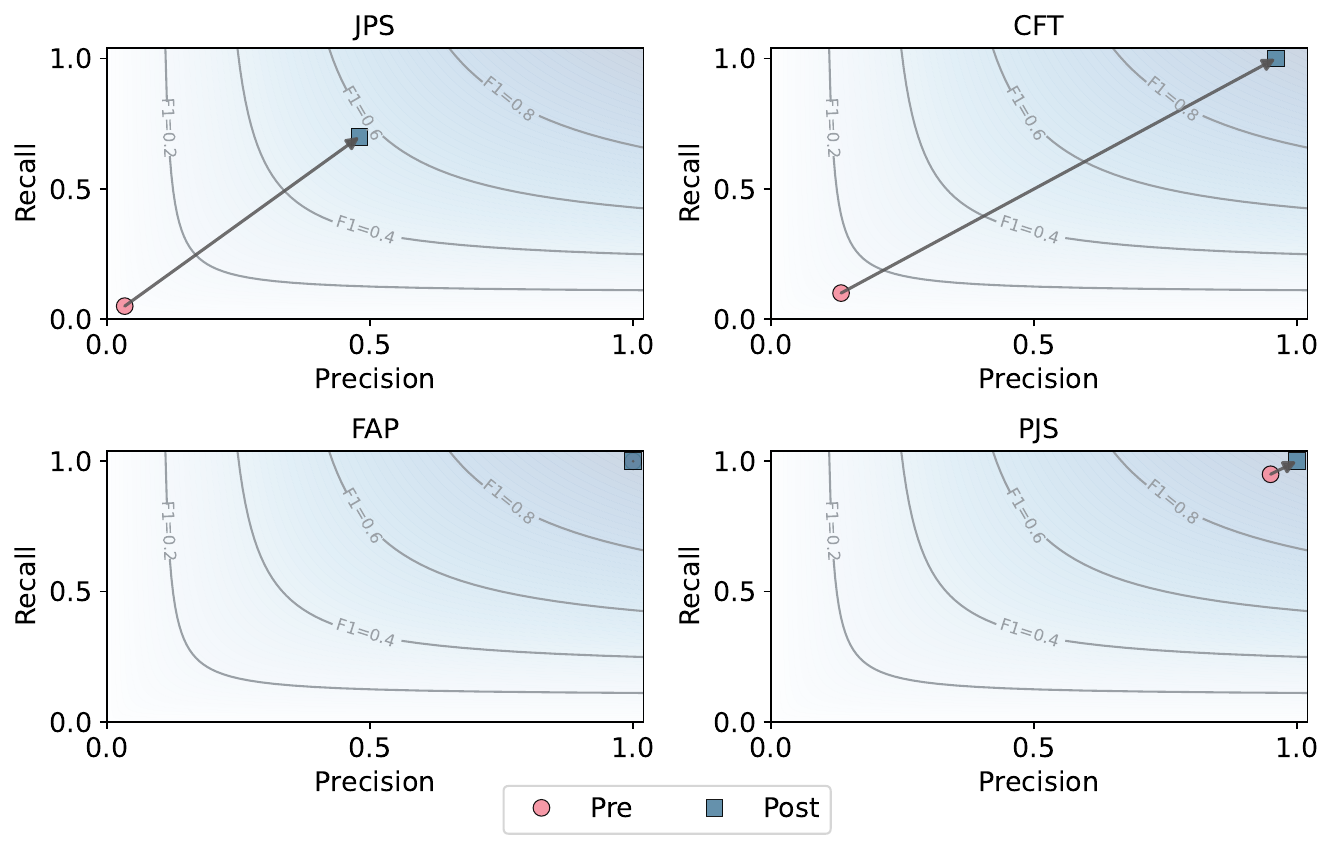}
    \caption{\textbf{Pre–post shifts in precision–recall space under {EndoRare}-based reader training.}
    Each panel corresponds to one entity (JPS, CFT, FAP, PJS). Circles mark the mean \emph{Pre} operating point across five readers and squares the mean \emph{Post} point; arrows connect \emph{Pre}\,$\rightarrow$\,\emph{Post} to indicate direction and magnitude of change.}
    \label{fig:education_curve}
  \end{minipage}
\end{figure}

To quantify the educational utility of EndoRare, we conducted a within-subject reader study assessing novice performance before and after exposure to synthetic cases (Fig.~\ref{fig:reader_training}a). The results reveal a clinically important pattern (Fig.~\ref{fig:reader_training}b).  For entities with distinctive phenotypes (FAP and PJS), novice endoscopists were already near ceiling at baseline, indicating that these conditions are readily recognizable even with limited exposure. In contrast, for cryptic entities such as JPS and CFT, which are morphologically ambiguous and can be mistaken for routine benign polyps, synthetic training produced large gains. Before training, novices frequently missed these lesions, leading to near-zero sensitivity. After exposure to diverse synthetic exemplars, performance increased substantially (Table~\ref{tab:reader_training_by_disease}). For JPS, precision increased from 0.033 to 0.480 ($\Delta=0.4467$, $p=0.0052$), with concurrent improvements in recall ($\Delta=0.6500$, $p=0.0004$). Gains were even larger for CFT, where recall reached 1.000 ($\Delta=0.9000$) alongside an increase in precision ($\Delta=0.8267$, $p=0.0004$). These results suggest that synthetic cases help trainees learn subtle visual cues that distinguish rare look-alike lesions from common findings.
\newpage
Importantly, improved sensitivity did not come at the cost of reduced precision. In the precision--recall plane (Fig.~\ref{fig:education_curve}), trajectories for JPS and CFT move upward and rightward, consistent with simultaneous gains in recall and precision. This pattern indicates that synthetic training does not simply encourage broader labeling of rare conditions, but improves discrimination.

Performance on the distinctive categories (FAP and PJS) remained stable without degradation ($p=0.3739$ for PJS), providing a safety signal that synthetic exposure does not induce confusion for conditions that are already visually recognizable. Together, these findings suggest that EndoRare can shorten the time needed to build diagnostic experience by providing concentrated exposure to rare, easily overlooked phenotypes. Full statistical details are provided in Supplementary Table~B7.

\subsection{Ablation Study on  EndoRare model architecture}

\begin{figure}[!b]
  \centering
  \includegraphics[width=\textwidth]{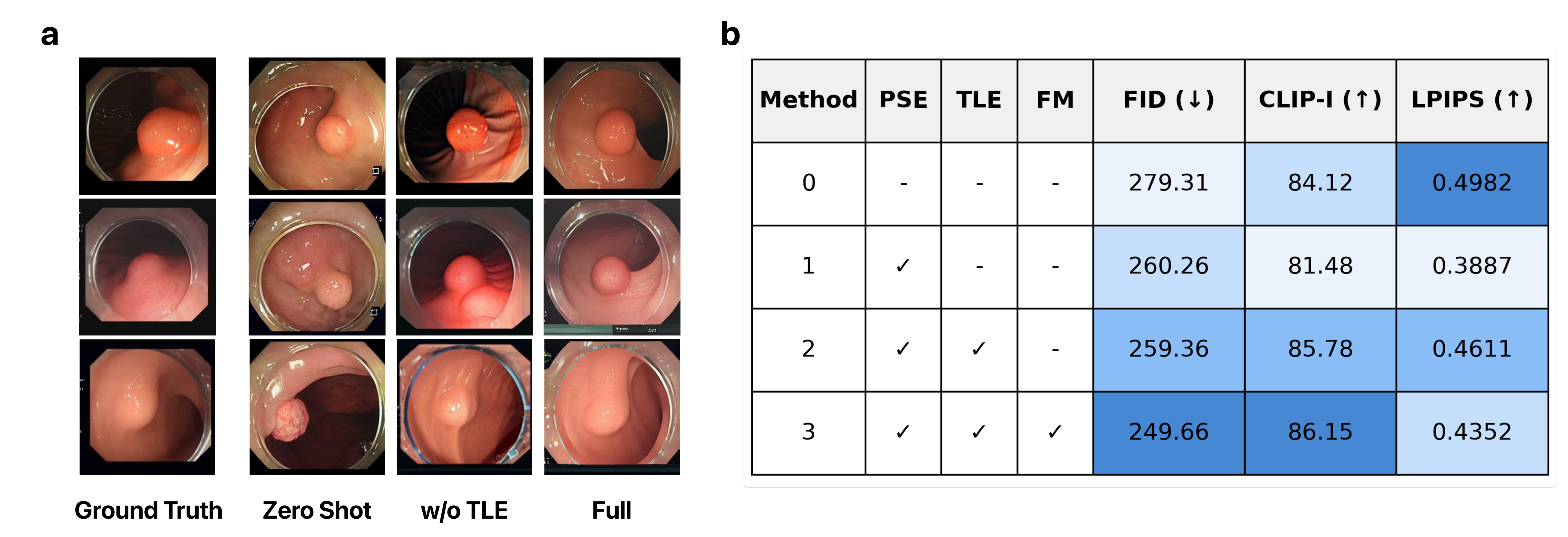}
  \caption{\textbf{Ablation of {EndoRare} components.}
  \textbf{a}, Qualitative comparison on representative endoscopic images: ground truth, a zero-shot baseline, a variant without the Tailored Lesion Embedding (w/o TLE), and the full model. 
  \textbf{b}, Quantitative ablation with modules toggled: Prototype-Specific Embedding (PSE), Tailored Lesion Embedding (TLE), and Fusion Module (FM). Lower FID and higher CLIP-I/LPIPS indicate better performance. The full model achieves the best overall balance.}
  \label{fig:ablation}
\end{figure}

To dissect the contribution of each architectural component, we performed an ablation in which PSE, TLE and FM were incrementally enabled and evaluated both qualitatively and quantitatively (Fig.~\ref{fig:ablation}a,b).

Qualitatively (Fig.~\ref{fig:ablation}a), the full model most closely resembles the reference examinations: lesions retain rounded contours and stalks in pedunculated polyps as well as the lobulated, multi-nodule appearance in clustered cases, while global colour tone and illumination are better matched. The zero-shot baseline tends to preserve only coarse class cues and drifts at the instance level (shape and surface texture). Removing TLE introduces visible artefacts (e.g., radial banding near the endoscope rim) and over-smoothed textures.

As shown in Fig.~\ref{fig:ablation}b, starting from the configuration without any module (FID\,=\,279.31; CLIP-I\,=\,84.12; LPIPS\,=\,0.4982), adding PSE reduces FID to 260.26 ($-19.05$; $-6.8\%$) but also lowers CLIP-I to 81.48 ($-2.64$) and LPIPS to 0.3887 ($-0.1095$). See Supplementary Fig.A15 for prototype reconstruction ability.
 Introducing TLE on top of PSE leaves FID essentially unchanged (259.36; $-$0.90) while substantially increasing CLIP-I to 85.78 (+4.30) and LPIPS to 0.4611 (+0.0724), indicating improved class alignment with greater diversity. Finally, enabling FM yields the best FID (249.66; $-$9.70 vs.\ w/o FM) and the highest CLIP-I (86.15; +0.37), with a small trade-off in LPIPS (0.4352; $-$0.0259 vs.\ w/o FM). Relative to the no-module baseline, the full model lowers FID by 29.65 ($-$10.6\%) and raises CLIP-I by 2.03 (+2.4\%), while maintaining competitive diversity (LPIPS 0.4352 vs.\ 0.4982).

\section{Discussion}\label{discussion}
Clinical endoscopy faces chronic data scarcity for rare entities: novice readers and supervised classifiers alike see too few examples to learn stable decision rules. Our goal was therefore pragmatic and clinically motivated: to supply realistic and class-faithful exemplars that broaden exposure without imposing additional collection burdens or privacy risks. We framed this as one-shot synthesis for rare lesions, with two key desiderata: fidelity to lesion identity and diversity along clinically meaningful axes (e.g., color, location). We evaluated performance not only with image-similarity metrics but also with task-level utility and clinician judgment~\cite{parasa2023framework}.

Conventional personalization by full fine-tuning (e.g., DreamBooth) can improve identity preservation but tends to shrink diversity and risks instance memorization under extreme data scarcity. {EndoRare} separates \emph{what} to keep (lesion identity) from \emph{what} to vary (non-critical attributes). Concretely, identity is recovered via prototype-based conditioning, and variability is injected by attribute-anchored text prompts, obviating any retraining of the diffusion model. This pipeline operates on prompts and prototypes rather than gradient updates over protected images, reducing both overfitting and privacy exposure relative to heavy-weight fine-tuning.

Our experiments show that {EndoRare} addresses the central challenge in image synthesis: balancing fidelity and diversity. As in Fig.~\ref{fig:case_study}b–c, it achieves substantially higher diversity (LPIPS $0.489\pm0.021$) while maintaining competitive fidelity (FID $280.26\pm11.64$) compared with established methods (Fine-tuned SD, DreamBooth), indicating avoidance of mode collapse without sacrificing anatomical realism. Its superior blinded clinical ratings (overall $2.32$) for realism and class faithfulness further support clinical plausibility; the close performance in semantic alignment (CLIP-I $0.847\pm0.006$) indicates strong image–image correspondence without degrading visual quality.

Downstream results corroborate these findings (Fig.~\ref{fig:classification}). Training with {EndoRare} augmentations improves PR-AUC ($0.708$ vs.\ $0.612$ for the strongest baseline) and ROC-AUC ($0.978$ vs.\ $0.961$), suggesting enhanced generalization rather than mere dataset inflation~\cite{dao2024application}. Gains are most pronounced where data are scarcest, as in CFT ($+0.100$), implying that EndoRare captures nuanced features of rare phenotypes that baselines miss.
 In reader studies, novices improved markedly on JPS (precision $0.033\!\rightarrow\!0.480$, F1 $0.040\!\rightarrow\!0.559$) and CFT (precision $0.133\!\rightarrow\!0.960$, F1 $0.114\!\rightarrow\!0.978$), indicating accelerated recognition of subtle pathological cues.

Because generic perceptual scores were developed for natural images, their relevance to endoscopy is limited: FID and related metrics can be insensitive to lesion-critical cues and rely on feature extractors misaligned with endoscopic content~\cite{deo2025metrics}. We therefore complemented them with blinded expert ratings as a clinical gold standard~\cite{liu2023observer} and, beyond image-level assessment, with task utility via classification and reader studies aligned to real-world endpoints~\cite{luo2025large}.

Recent work suggests that diffusion models, especially when fine-tuned with weak regularization, can memorize patient images and leak prototypes upon querying~\cite{carlini2023extracting,dar2025unconditional,chen2024towards,somepalli2023understanding}.
In our experiments, instance-level fine-tuned baselines (Fintune SD and DreamBooth) exhibited lower diversity at comparable fidelity~\cite{chen2023disenbooth}, consistent with this risk profile. {EndoRare} mitigates memorization by avoiding gradient updates on patient pixels, enforcing a language-informed factorization, and sampling with attribute variation that discourages verbatim replication. Mitigation, however, is not proof of absence; future releases should include formal privacy audits (nearest-neighbour and attribute-leakage analyses, membership/attribute-inference tests), configurable safety margins (e.g., minimum attribute perturbations), and optional safeguards such as watermarking or differentially private training for auxiliary adapters~\cite{liu2023observer,hur2024latent,daum2024differentially}.

This study has limitations. We evaluate four entities with five novice readers; larger multi-center studies across devices and expertise levels are needed to assess durability and generalizability~\cite{yang2024limits}. Ceiling effects for some categories (FAP, PJS) limit detectable gains; future work should emphasize harder phenotypes and mixed-prevalence test sets to probe sensitivity–specificity trade-offs. Although clinical ratings and task-level endpoints improve interpretability beyond FID, the field lacks community-accepted, lesion-aware metrics (e.g., pathology-consistent feature tests, segmentation- or report-aligned consistency). The approach also assumes high-quality textual descriptors; robust report parsing and attribute validation from routine endoscopy reports are important next steps. Finally, while {EndoRare} reduces privacy exposure relative to fine-tuning, it does not eliminate it; standardizing privacy tests and red-team prompts in the release process should become required practice.

Taken together, carefully controlled, retraining-free synthetic augmentation can expand exposure to rare phenotypes for both humans and algorithms, improve classifier operating points at low false-positive rates, and offer a safer alternative to heavy fine-tuning that curbs instance memorization~\cite{yao2024risks}. By prioritizing expert assessment and task utility alongside generic image scores, this work aligns generation with clinical value. In the near term, {EndoRare} can serve as a companion to reader education and model development in low-prevalence settings; in the longer term, the same disentanglement-and-control principles could support case-mix simulation, curriculum learning, and bias probing across institutions. Progress will hinge on standardizing clinical evaluations, auditing privacy systematically, and extending controllable generation to a broader spectrum of rare diseases and acquisition settings.

\section{Methods}\label{method}
\subsection{Problem Formulation and Preliminaries}

\noindent\textbf{Task Definition.}  
We define the task of \emph{one-shot rare disease image synthesis} as follows. Let $\mathcal{D}_S = \{(I_{S_i}, R_{S_i})\}_{i=1}^{N_S}$ be source dataset of common disease images, where each image $I_{S_i} \in \mathbb{R}^{H \times W \times 3}$ denotes an RGB endoscopy image and $R_{S_i} \in \mathcal{R}$ is its associated text report. Notably, such source datasets are readily compilable from clinical records ~\cite{11080481}.
Given a single target image of rare disease $\mathcal{D}_T = \{I_T\}$, our goal is to generate a synthetic set $\mathcal{D}_G = \{I_G^j\}_{j=1}^{N_G}$ that preserves the unique characteristics of $I_T$ while ensuring sufficient fidelity and diversity.

\noindent\textbf{Diffusion Models.}  
Diffusion Models (DMs)~\cite{ho2020denoising} generate images by progressively denoising a noisy sequence. Given an image $I_0$, Gaussian noise is added incrementally, and a U-Net~\cite{ronneberger2015u} $\epsilon_\theta(\cdot, t)$ is trained to predict $\epsilon_t$ at timestep $t$. Conditional generation incorporates text embeddings from a pretrained CLIP encoder $\tau_\theta(y)$~\cite{radford2021learning} via cross-attention (see Supplementary Fig.A14). We adopt Stable Diffusion~\cite{rombach2022high}, a Latent Diffusion Model (LDM) that operates in latent space for efficiency. The image is encoded as \( z = \mathcal{E}(I_0) \) using a Variational Autoencoder~\cite{kingma2013auto}, and the conditional LDM training objective is:
\begin{equation}
\label{loss_DM}
    \mathcal{L}_{\text{LDM}} 
    = \mathbb{E}_{\textbf{z}, y, \epsilon \sim \mathcal{N}(0,1), t} 
    \left[ \left\| \epsilon_t - \epsilon_\theta(\textbf{z}_t, t, \tau_\theta(y)) \right\|^2 \right],
\end{equation}
where $\mathbf{z}_t$ is the noised latent at timestep $t$ and $y$ is a text condition.

\noindent

\subsection{Model Overview}
\begin{figure}[t]
    \centering
    \includegraphics[width=\textwidth]{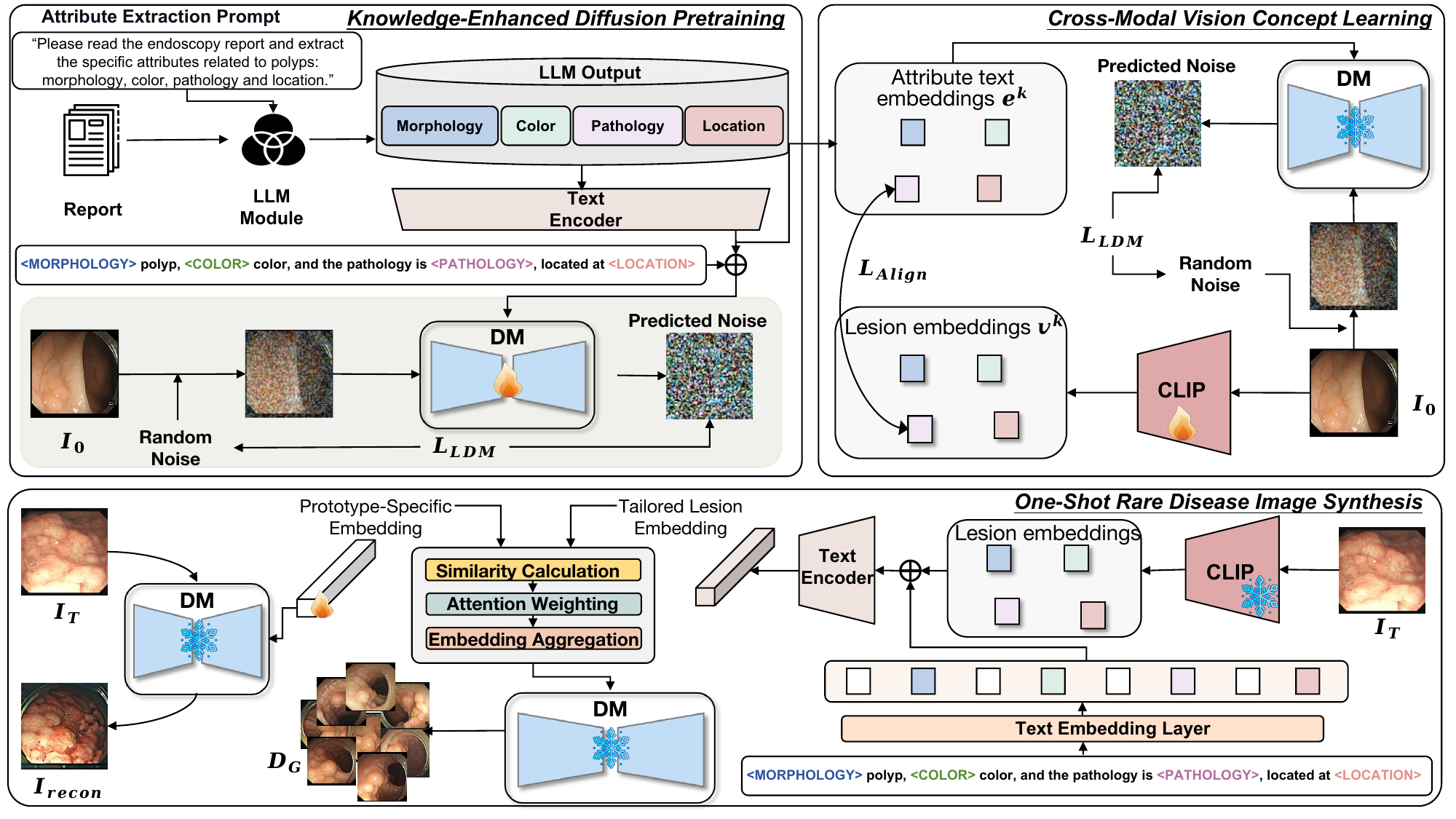}
    \caption{Framework of {EndoRare} for one-shot disease image synthesis in endoscopy.}
    \label{Framework}
\end{figure}

As illustrated in Fig.~\ref{Framework}, we first pretrain a text-to-image diffusion model on source datasets of common diseases. To steer learning toward clinically meaningful features, senior attending endoscopists re-annotate and enrich the original endoscopy reports, after which a large language model extracts salient attributes—morphology, color, pathology, and location—from these curated texts. We reorganize the extracted attributes into compositional prompts that provide a structured prior over typical lesion characteristics. Next, we introduce a cross-modal lesion–feature disentanglement module (Sec.~\ref{encoder}) to align visual factors with their textual counterparts. After this concept-learning stage, given a single prototype image of a rare disease, we translate its disentangled visual factors into textual descriptions and combine them with learnable, prototype-specific embeddings to synthesize the final image.

\subsection{Knowledge-Enhanced Diffusion Pretraining}
\label{3.2}

To integrate domain knowledge from endoscopy reports in $\mathcal{R}$, we use a Large Language Model (LLM) to extract four key attributes: morphology, color, pathology, and location. These attributes are formatted into a structured template: a \texttt{<MORPHOLOGY>} polyp, \texttt{<COLOR>} color, and the pathology is \texttt{<PATHOLOGY>}, located at \texttt{<LOCATION>}. The formatted text is encoded by a pretrained CLIP text encoder to obtain $\tau_\theta(y_i)$, which conditions the diffusion model.  
By optimizing the loss in Eq.~\eqref{loss_DM} with these structured inputs, we embed domain-specific knowledge, enabling the model to capture meaningful lesion characteristics. This pretraining provides a broad representation of common disease appearances, ensuring generalization to rare lesions.

\subsection{Cross-Modal Visual Concept Learning}
\label{encoder}

To disentangle intricate visual features of rare disease images, we introduce a cross-modal lesion feature disentanglement mechanism. This framework aligns attribute-specific visual representations of \( I_T \) with their corresponding textual embeddings.
For each endoscopy report \( R_{S_i} \in \mathcal{D}_S \), we extract key textual attributes, denoted as \( \mathit{Attr_i} \). Each attribute \( attr_i^k \in \mathit{Attr_i} \) is encoded using the CLIP text encoder \( \tau_\theta \), producing an embedding \( \mathbf{e}_i^k = \tau_\theta(attr_i^k) \). Simultaneously, a dedicated branch processes the corresponding image \( I_{S_i} \) using a CLIP image encoder to obtain \( \mathbf{v}_i^k \). 
To enforce cross-modal consistency, we integrate an attribute alignment term into the LDM loss, optimizing the following objective:
\begin{equation}
    \mathcal{L}_{\text{total}} = \mathbb{E}_{\mathcal{D}_S, \epsilon \sim \mathcal{N}(0,1), t} \left[ \sum_{k} \left\| \mathbf{v}^k - \mathbf{e}^k \right\|^2 + \lambda \left\| \epsilon_t - \epsilon_\theta(\mathbf{z}_t, t, \tau_\theta(y)) \right\|^2 \right],
    \label{eq:total_loss}
\end{equation}
where \( \lambda \) balances attribute alignment with the diffusion model’s reconstruction loss. Here, \( y \) is the formatted textual description of \( R_{S} \) as defined in Section~\ref{3.2}.

\subsection{One-Shot Rare Disease Image Synthesis}
\label{generate}

\textbf{Prototype-Specific Embedding.}  
To generate rare disease images without retraining the diffusion model, we introduce a learnable vector \( \mathbf{p}_T \), the \emph{Prototype-Specific Embedding}, which captures the unique features of a target lesion \( I_T \). Given the noised latent sample \(\mathbf{z}_t\) at timestep \( t \), the model \(\epsilon_\theta\) is conditioned on \( \mathbf{p}_T \) to predict the noise. The Mean Squared Error (MSE) loss optimizes \( \mathbf{p}_T \) by comparing the predicted noise \(\epsilon_\theta(\mathbf{z}_t, t, \mathbf{p}_T)\) with the actual noise \(\epsilon_t\):
\begin{equation}
\mathcal{L}_{\text{Recon}} 
= \mathbb{E}_{\mathbf{z},\mathbf{p},\epsilon \sim \mathcal{N}(0,1), t} 
\left[ 
\left\| \epsilon_t - \epsilon_\theta\bigl(\mathbf{z}_t, t, \mathbf{p}_T\bigr)\right\|^2 
\right].
\end{equation}
With the pretrained diffusion model \(\epsilon_\theta\) fixed, adaptation occurs only in \( \mathbf{p}_T \). To provide effective guidance, \( \mathbf{p}_T \) is initialized with the text embedding of the lesion class name from \(\tau_\theta\).  

\noindent\textbf{Tailored Lesion Embedding.}  
While \( \mathbf{p}_T \) reconstructs \( I_T \), the generated image \( I_{\text{recon}} \) may lack diversity and realism. To address this, we leverage the cross-modal visual concept learning, where each CLIP-based encoder extracts attribute embeddings \(\{\mathbf{v}_T^k\}\) from \( I_T \). These embeddings are formatted into the textual template (Section~\ref{3.2}), forming the \emph{Tailored Lesion Embedding} \( \mathbf{r}_T \). 

\noindent\textbf{Embedding Fusion with Guided Denoising.}  
To generate high-fidelity rare disease images, we fuse \( \mathbf{p}_T \) with \( \mathbf{r}_T \), where \( \mathbf{p}_T \) retains core morphology and \( \mathbf{r}_T \) enhances diversity through attribute-based variations. The fusion process involves computing similarity matrices between \( \mathbf{r}_T \) and \( \mathbf{p}_T \), deriving attention weights via softmax, and aggregating embeddings into the final fused representation \( \mathbf{e}_T \), either by averaging or concatenation.
The fused embedding \( \mathbf{e}_T \) then guides the denoising process:
\begin{equation}
\begin{aligned}
\mathbf{z}_{t-1} = & \sqrt{\bar{\alpha}_{t-1}} \left( \frac{\mathbf{z}_t - \sqrt{1 - \bar{\alpha}_t} \epsilon_{\theta}(\mathbf{z}_t, t, \mathbf{e}_T)}{\sqrt{\bar{\alpha}_t}} \right) 
 + \sqrt{1 - \bar{\alpha}_{t-1} - \sigma^2} \epsilon_{\theta}(\mathbf{z}_t, t, \mathbf{e}_T) + \sigma \epsilon,
\end{aligned}
\end{equation}
where \(\epsilon_{\theta}\) predicts noise given \(\mathbf{z}_t\), \(\mathbf{e}_T\), and timestep \(t\). This iterative process continues until the fully denoised latent code \(\mathbf{z}_0\) is obtained and decoded by \(\mathcal{D}(\mathbf{z}_0)\) into the final endoscopic image.

\section{Data availability}
Restrictions apply to the availability of the developmental and validation datasets. 
These datasets were used with participants' permission for the present study and are not publicly available. 
De-identified data may be made available for research purposes from the corresponding authors upon reasonable request 
and subject to institutional and ethical approvals.
\section{Code availability}
The code to reproduce the results is available at
\href{https://github.com/Jia7878/EndoRare}{\texttt{github.com/Jia7878/EndoRare}}.


\end{document}